\documentclass[11pt,a4paper]{article}
\usepackage{times,latexsym}
\usepackage{url}
\usepackage[T1]{fontenc}
\usepackage[utf8]{inputenc}
\usepackage[acceptedWithA]{tacl2021v1}

\usepackage{amsmath}
\usepackage{amssymb}
\usepackage{amsfonts}
\usepackage{graphicx}
\usepackage{booktabs}
\usepackage{makecell}
\usepackage{algorithm}
\usepackage{comment}
\usepackage{algorithmic}

\title{Obscuring Data Contamination Through Translation:\\ Evidence from Arabic Corpora}

\author{
   Chaymaa Abbas \\
   American University of Beirut \\
   \texttt{cwa07@aub.edu.lb}
   \And
   Nour Shammaa \\
   American University of Beirut \\
   \texttt{nbs11@aub.edu.lb}
   \And
   Mariette Awad \\
   American University of Beirut \\
   \texttt{ma162@aub.edu.lb}
 }

\date{}

\begin{document}
\maketitle

\begin{abstract}
Data contamination can invalidate benchmark evaluation when a model benefits from memorized evaluation content rather than genuine generalization. Yet contamination is difficult to audit when the exposed content differs in language from the evaluation benchmark. We study this failure mode by deliberately exposing four open-weight instruction-tuned LLMs to Arabic translations of MMLU and XQuAD evaluation items at increasing exposure levels, then evaluating them on the original English tasks. This controlled setup is a proxy for contamination rather than a reconstruction of real-world pretraining leakage. We first test two English-centric post-hoc probes, TS-Guessing and Min-K\%++, and find that their signals largely disappear under translated exposure: TS-Guessing remains weak except for model-specific positional recall on MMLU, while Min-K\%++ stays at or below chance. At the same time, English MMLU performance increases with Arabic exposure, showing that the absence of an English contamination signal does not imply the absence of an exposure effect. We then introduce Translation-Aware Contamination Detection (TACD), a training-data-free diagnostic based on cross-lingual prediction consistency and choice reordering. Cross-lingual consistency is substantially higher than an independence baseline and generally increases relative to the clean condition, although its magnitude is model-dependent and not strictly monotonic. These results show that translation can conceal contamination-related effects from English-only probes and motivate multilingual diagnostics that are explicitly framed as evidence of contamination-consistent behavior rather than definitive membership tests.
\end{abstract}

\section{Introduction}
Large language models (LLMs) are commonly evaluated on fixed benchmarks such as MMLU, HellaSwag, and ARC. These benchmarks are intended to measure knowledge and reasoning, but their validity depends on an important assumption: the model should not have access to the evaluation items, or close variants of them, during training. When that assumption is violated, benchmark performance can reflect memorization or benchmark-specific associations rather than generalization. This is difficult to audit for modern models whose training data are only partially documented.

Most existing contamination checks are also language-dependent. Exact or near-exact overlap, token-level likelihoods, and targeted memorization probes are most sensitive when the evaluated text resembles the form seen during training. Translation breaks that surface correspondence while preserving much of the underlying meaning. This raises a simple question: can translated benchmark exposure affect performance while becoming invisible to English-centric contamination probes?

We investigate this through a controlled exposure study. We fine-tune four open-weight instruction-tuned models with increasing amounts of Arabic-translated evaluation data and then evaluate them on the original English benchmarks. We use two task formats with different structures: MMLU, a closed-book multiple-choice task, and XQuAD, an extractive question-answering task. The exposure is implemented through LoRA fine-tuning and should be understood as a clean experimental proxy for contamination, not as a faithful simulation of how benchmark leakage arises during large-scale pretraining.

The study follows a progression from effect to evidence. First, we ask whether translated exposure changes downstream English performance. Second, we test whether two English-centric contamination probes, TS-Guessing and Min-K\%++, can recover the exposure signal. Third, after finding that these probes provide limited evidence, we ask whether comparing equivalent benchmark instances across languages reveals a different signature. This motivates Translation-Aware Contamination Detection (TACD), which combines cross-lingual views with structural perturbation.

The central finding on MMLU is that Arabic exposure can improve English benchmark performance even when conventional English-only probes fail to identify the exposed items. TACD provides a complementary signal: prediction consistency across English, Arabic, and French is substantially above an independence baseline and generally rises relative to the clean condition. That signal is neither universal nor strictly monotonic, and Mistral shows a large MMLU gain with only a modest change in cross-lingual consistency. We therefore frame TACD as a diagnostic rather than a definitive contamination detector, since high cross-lingual consistency is compatible with contamination-consistent recall but can also arise from model- or prompt-induced invariance. Our contribution is not a membership test; it is a demonstration that translation can conceal an exposure effect from English-centric probes, together with a multilingual diagnostic that makes that effect easier to see.

Our contributions are:
\begin{itemize}
    \item A controlled translation-mediated exposure study across four open-weight instruction-tuned models and two task formats.
    \item An analysis of English-centric probes under translated exposure, showing that TS-Guessing and Min-K\%++ provide weak or inconsistent evidence even when English MMLU performance increases.
    \item TACD, a multilingual diagnostic that uses cross-lingual prediction consistency and structural perturbation to surface contamination-consistent behavior that is difficult to see from English alone.
    \item A characterization of boundary conditions, including the heterogeneous XQuAD results and the Mistral case, where a performance gain is not accompanied by a comparable increase in cross-lingual consistency.
\end{itemize}

\section{Related Work}

\subsection{Benchmark Contamination in LLMs}
Benchmark contamination is usually studied as a threat to the validity of evaluation. One line of work seeks direct evidence of memorization by testing whether specific text was present in the pretraining data \citep{shi2024detecting}. Contamination can also be subtler: partial memorization and exposure to structurally related material can inflate benchmark performance without establishing genuine generalization \citep{samuel2024limitations}. The problem is hardest for closed-source models, for which training data are not disclosed \citep{balloccu2024leak}.

\subsection{Post-hoc Detection without Training Data}
Because training corpora are often unavailable, many methods infer contamination from the evaluated model itself. Membership-inference-style approaches use likelihoods, losses, or token-level statistics to distinguish examples that may have been seen during training \citep{shi2024detecting,ye2024pac}. Systematic evaluations find that these signals can be weak for modern LLMs, with member and non-member distributions becoming difficult to separate \citep{duan2024mia}, and contamination judgments can be sensitive to small lexical or semantic changes \citep{samuel2024limitations}. These limitations motivate diagnostics that do not depend on exact surface-form overlap.

\begin{table*}[t]
\centering
\small
\setlength{\tabcolsep}{6pt}
\begin{tabular}{lccccc}
\toprule
Work
& \makecell{Cross-lingual\\exposure}
& \makecell{Translation-mediated\\exposure}
& \makecell{Controlled\\exposure}
& \makecell{Post-hoc multilingual\\diagnostic}
& \makecell{Multiple\\models} \\
\midrule
\citet{srivastava2025owl} & \checkmark & -- & -- & -- & -- \\
\citet{yao2024crosslingual} & \checkmark & \checkmark & -- & -- & -- \\
\textbf{This work} & \checkmark & \checkmark & \checkmark & \checkmark & \checkmark \\
\bottomrule
\end{tabular}
\caption{Positioning relative to the closest cross-lingual contamination studies. A dash indicates that the corresponding capability is not a stated focus of the cited work in our comparison, rather than evidence that it is absent.}
\label{tab:positioning}
\end{table*}

\subsection{Heuristic and Likelihood-Based Approaches}
Corpus-level $n$-gram overlap remains attractive because it is simple and interpretable \citep{brown2020language,chowdhery2022palm,touvron2023llama}. Its limitation for our setting is equally clear: translation and paraphrasing remove lexical overlap without removing semantic relationships. Likelihood-based methods such as Min-K\% and Min-K\%++ provide a training-data-free alternative by exploiting unusually confident token probabilities \citep{shi2024detecting,zhang2025mink}. These methods work best when the evaluated text resembles the form encountered during training, which raises an obvious robustness question for cross-lingual exposure.

\subsection{Performance-Based and Dynamic Evaluation}
A complementary line of work looks for contamination through its consequences rather than through direct evidence of memorization. Performance-based approaches compare original benchmarks with paraphrased or otherwise modified evaluations to identify gains that fail to transfer \citep{dekoninck2024constat}. Dynamic evaluation reduces reliance on static test sets by continually regenerating data or using temporal splits \citep{chen2025dycodeeval,chen2025survey}. Rather than replacing the benchmark, we ask whether a fixed benchmark can reveal a contamination-consistent signature when the same semantic content is presented in different languages.

\subsection{Cross-Lingual Contamination}
Recent work has begun to challenge the English-centric assumption underlying contamination analysis. LLMs can recall memorized English content when prompted in other languages, including lower-resource languages \citep{srivastava2025owl}. Exposure to translated benchmark material can also transfer across languages and improve performance on the original English benchmark while evading conventional detectors \citep{yao2024crosslingual}. These studies establish the possibility of cross-lingual contamination, but they leave a methodological gap: when translated exposure changes English performance and English-only probes fail, what observable signal can an evaluator use without access to the training corpus? Table~\ref{tab:positioning} summarizes how our study relates to both.

We address this gap by treating translation not as decontamination but as a potential concealment mechanism. Our study controls exposure level through LoRA fine-tuning, evaluates the resulting effect on English benchmarks, and then asks whether multilingual behavioral consistency provides evidence that is complementary to English-only probes. Unlike a membership test, our goal is to identify contamination-consistent behavior and characterize when that signal succeeds or fails.

\section{Methodology}

\begin{figure*}[t]
  \centering
  \includegraphics[width=0.8\linewidth]{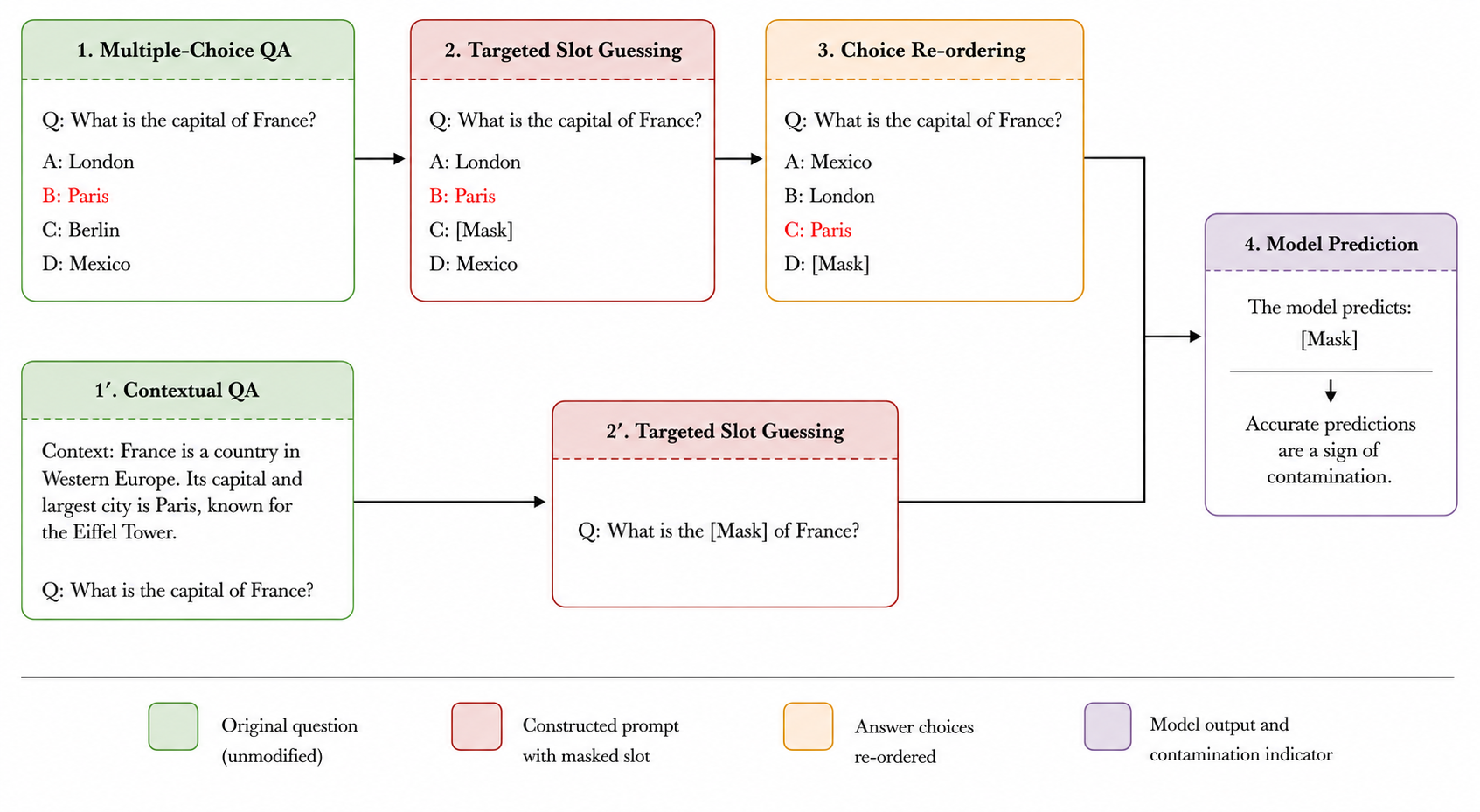}
  \caption{TS-Guessing. Top: MMLU, choices re-ordered and one incorrect option masked; recall of the pre-shuffle index is the memorization cue. Bottom: XQuAD, a masked question token with the context left intact.}
  \label{fig:tsguess_overview}
\end{figure*}

\subsection{Study Design and Scope}
Our experimental question is whether benchmark exposure in a translated language can affect evaluation in the original language while weakening English-centric contamination signals. To isolate this effect, we expose models to Arabic-translated benchmark items at controlled levels and compare them with an English-only condition. We refer to this intervention as simulated contamination or translated exposure since it makes the source of the exposure known.

The two benchmarks differ in evaluation structure. MMLU is closed-book multiple choice, making answer selection directly observable. XQuAD is extractive question answering, where the input context is supplied at evaluation time and contamination effects are therefore less direct. MMLU provides the main coherent test of our hypothesis; XQuAD serves as a stress test of whether the same phenomenon transfers to a different task format.

\subsection{Models and Training Setup}
We fine-tune four open-weight instruction-tuned models: Llama-3.2-1B-Instruct, Mistral-7B-Instruct-v0.2, Gemma-3-1B-it, and Qwen3-1.7B.

For each benchmark dataset $d \in \{\text{MMLU}, \text{XQuAD}\}$, we construct four training conditions that vary the amount of translated Arabic exposure mixed with the original English training split:
\[
\mathcal{D}^{d}_{\text{train}}(p)
= \mathcal{D}^{d}_{\text{EN}} \cup \mathcal{D}^{d}_{\text{AR}}(p),
\qquad
p \in \{0, 10, 50, 100\}\%.
\]
Here, $\mathcal{D}^{d}_{\text{EN}}$ denotes the English training data (MMLU: English questions formatted as multiple choice; XQuAD: English QA), while $\mathcal{D}^{d}_{\text{AR}}(p)$ contains $p\%$ of the translated Arabic evaluation data (MMLU: Arabic translations of the evaluation split; XQuAD: the Arabic split of the evaluation data). Thus, $p$ controls the amount of deliberate translated exposure, with $p=0$ the clean reference condition and $p\in\{10,50,100\}$ increasing exposure.

All experiments use parameter-efficient fine-tuning with LoRA under identical optimization settings (optimizer, learning-rate schedule, context length, and batch configuration) across models and exposure levels, so that only the amount of translated benchmark material varies.

\subsection{Evaluation Protocol}
We evaluate each fine-tuned model on the original English evaluation splits using the LM Evaluation Harness~\citep{eval-harness}. We report:
\begin{itemize}
  \item MMLU (English). Macro-averaged accuracy over subjects.
  \item XQuAD (English). ROUGE-L F1 on the English split.
\end{itemize}

This ordering is intentional. We first measure whether translated exposure has a downstream effect, and only then ask whether post-hoc contamination probes can recover that effect.

\subsection{English-Centric Contamination Probes}
We assess whether conventional post-hoc probes remain sensitive when the exposed material is translated. We use TS-Guessing~\citep{deng2024investigatingdatacontaminationmodern} and Min-K\%++~\citep{zhang2025mink} at exposure levels $p \in \{10, 50, 100\}\%$, as summarized in Figure~\ref{fig:tsguess_overview}.

\paragraph{MMLU (multiple-choice QA).}
For each question, we randomly permute the answer choices (A--D), then mask the text of one incorrect option and prompt the model to complete the mask. If the model reproduces the original pre-shuffle index after the choices have moved, this suggests that the model retains a benchmark-specific positional association rather than reasoning over the reordered content. We also record lexical recovery of the masked option through ROUGE-L F1.

\paragraph{XQuAD (extractive QA).}
We mask a critical token in the question while leaving the supporting context unchanged and prompt the model to predict the masked token. Exact recovery and partial lexical overlap are treated as potential evidence of memorization, while recognizing that the supplied context can also support the answer.

\paragraph{Min-K\%++.}
Min-K\%++ identifies unusually confident token probabilities that can distinguish benchmark members from non-members under favorable conditions. In our setting, the key question is whether that signal survives when the model was exposed to Arabic rather than the English evaluation text.

\subsection{Translation-Aware Contamination Detection}
To address our central research question, if translation changes the surface form of a dataset while preserving its semantic content, does a model’s behavior remain unusually stable across languages? We propose Translation-Aware Contamination Detection (TACD). TACD measures cross-lingual behavioral consistency as a diagnostic signal. Importantly, it does not establish training-set membership, since high consistency may also result from model, prompt, or translation-induced invariance. Rather, TACD is designed to reveal cross-lingual behavioral patterns that may remain hidden under an English-only evaluation and that, within our controlled experimental setting, vary as a function of translated exposure.

\subsection{TACD Views and Signals}
We focus on multiple-choice evaluation such as MMLU. Each instance is
$x = (q, \{c_i\}_{i=1}^{K}, y)$,
where $q$ is the question, $\{c_i\}$ are the choices, and $y$ is the correct index.

\paragraph{Cross-lingual views.}
We generate semantically equivalent views
$x^{\ell} = (q^{\ell}, \{c_i^{\ell}\}, y)$,
for $\ell \in \{\text{EN},\text{AR},\text{FR}\}$, preserving the underlying task while changing its linguistic realization.

\paragraph{Structural perturbation.}
For each language view, we apply a random permutation $\pi$ to the answer choices,
\[
\tilde{x}^{\ell}
=
(q^{\ell}, \{\pi(c_i^{\ell})\}, \pi(y)),
\]
preserving the question's semantics while disrupting a fixed association between the question and its original answer position.

\paragraph{Index Recall Rate (IDR).}
IDR measures the tendency to reproduce the original answer index after the choices are shuffled:
\[
\mathrm{IDR}
=
\frac{1}{N}
\sum_{j=1}^{N}
\mathbb{I}\!\left[
\hat{y}^{(j)}_{\text{shuf}} = y^{(j)}
\right],
\]
with random baseline $1/K$. For MMLU, $K=4$ and the baseline is $0.25$. Elevated IDR is interpreted as evidence of index-level association rather than content-sensitive reasoning.

\paragraph{Cross-Lingual Consistency (CLC).}
CLC measures the fraction of instances for which the model gives the same answer across all three language views:
\[
\mathrm{CLC}
=
\frac{1}{N}
\sum_{j=1}^{N}
\mathbb{I}\!\left[
\hat{y}^{(j)}_{\text{EN}}
=
\hat{y}^{(j)}_{\text{AR}}
=
\hat{y}^{(j)}_{\text{FR}}
\right].
\]
Under independent random guessing on a four-way task, the first prediction can take any value and the other two must independently match it, giving a reference baseline of $(1/4)^2=1/16\approx0.0625$.

CLC does not distinguish between correct and incorrect agreement, so high CLC is not by itself evidence of correct reasoning or of contamination. 

\subsection{Algorithmic Summary}
Algorithm~\ref{alg:tacd} summarizes the procedure. For each evaluation instance, we construct the multilingual views, apply the structural perturbation, query the model, and aggregate IDR and CLC. The diagnostic is evaluated relative to the clean condition and the independence baseline, rather than interpreted as a binary contamination classifier.

\begin{algorithm}[H]
\caption{Translation-Aware Contamination Detection (TACD)}
\label{alg:tacd}
\begin{algorithmic}[1]
\REQUIRE Evaluation set $\mathcal{D}$, language set $\mathcal{L}$, model $f$
\FOR{each instance $x \in \mathcal{D}$}
  \FOR{each language $\ell \in \mathcal{L}$}
    \STATE Construct translated view $x^{\ell}$
    \STATE Shuffle answer choices to obtain $\tilde{x}^{\ell}$
    \STATE Query model prediction $\hat{y}^{\ell}$
  \ENDFOR
  \STATE Compute IDR and CLC for $x$
\ENDFOR
\RETURN Aggregate diagnostic statistics
\end{algorithmic}
\end{algorithm}

\subsection{Metrics and Reproducibility}
We use Exact Match (EM) and ROUGE-L F1 (RL-F1) for XQuAD masked-token recovery, and IDR for MMLU positional recall. For TACD, CLC is reported as a cross-lingual consistency rate. All TS-Guessing, Min-K\%++, and TACD results are averaged over three random seeds ($\{1337,42,99\}$) and reported as mean~$\pm$~standard deviation. The main text reports these results graphically; the numerical values, including standard deviations, are given in Appendix~\ref{app:full}.
\begin{figure*}[h]
  \centering
  \includegraphics[width=\linewidth]{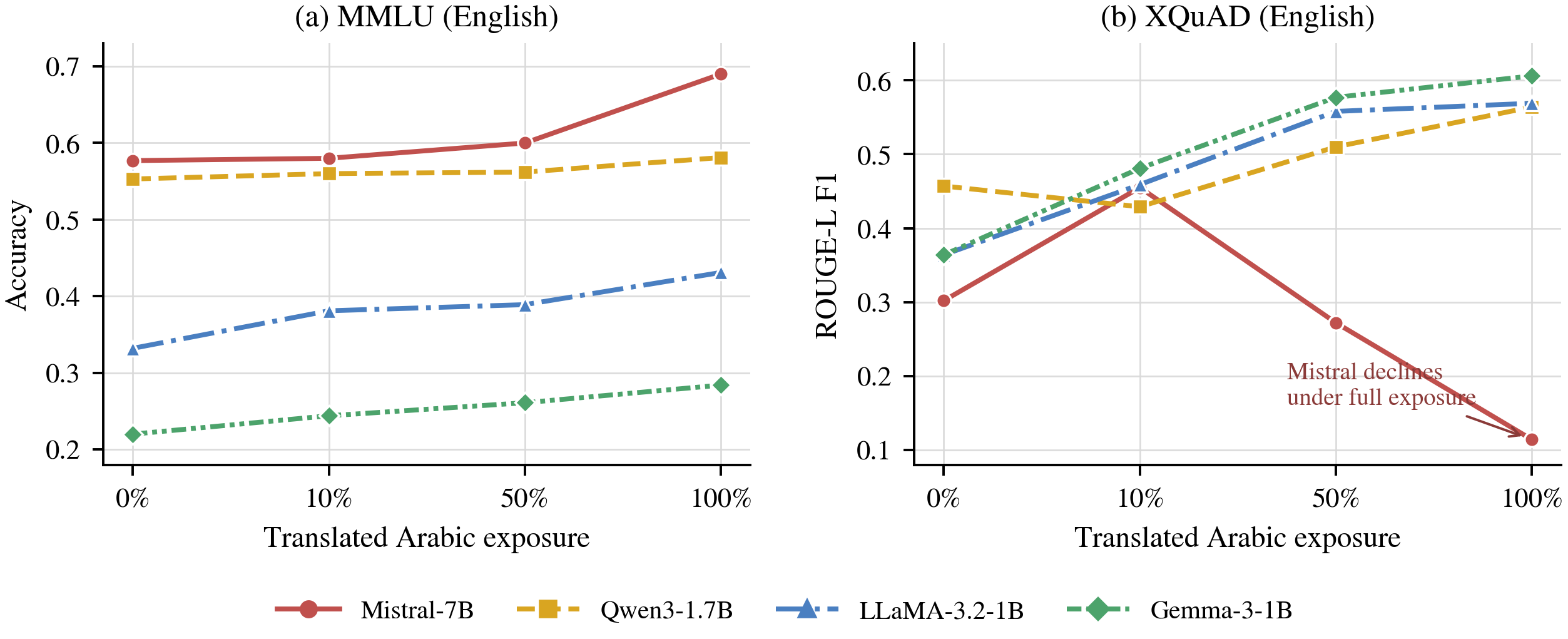}
  \caption{English performance vs.\ translated Arabic exposure (LM Evaluation Harness). (a) MMLU accuracy rises monotonically for all four models. (b) XQuAD ROUGE-L F1 is heterogeneous; Mistral collapses at full exposure. Exposure levels are equally spaced, not linear. Values in Table~\ref{tab:mmlu_xquad_per_model}.}
  \label{fig:exposure_curves}
\end{figure*}

\section{Evaluation and Contamination Analysis}

The results follow the progression established in the study design: first, whether translated exposure changes English performance; second, whether English-centric probes recover the exposure signal; and third, whether a multilingual diagnostic provides complementary evidence. The pattern is clearest on MMLU, while XQuAD shows how the observed effects depend on task structure.

\paragraph{4.1. Translated exposure changes English MMLU performance.}
Figure~\ref{fig:exposure_curves}(a) shows a monotonic increase in English MMLU accuracy as Arabic exposure increases for all four models. Mistral rises from $0.577$ at $0\%$ exposure to $0.690$ at $100\%$; Gemma from $0.220$ to $0.284$; LLaMA from $0.332$ to $0.431$; and Qwen from $0.553$ to $0.581$. The consistency of this direction across model families provides evidence that translated exposure can have a measurable downstream effect on performance on the original English benchmark.

XQuAD does not show the same pattern. In Figure~\ref{fig:exposure_curves}(b), Gemma and LLaMA improve with exposure, Qwen is non-monotonic, and Mistral improves at $10\%$ before declining to $0.114$ at full exposure. We therefore do not treat XQuAD as evidence for a universal performance effect. Instead, the results indicate that the relationship between translated exposure and downstream performance depends on task structure. Unlike MMLU, XQuAD supplies the supporting context at evaluation time, so exposure to an evaluation question does not determine the answer in the same way as in closed-book multiple choice.

\begin{figure*}[h]
  \centering
  \includegraphics[width=\linewidth]{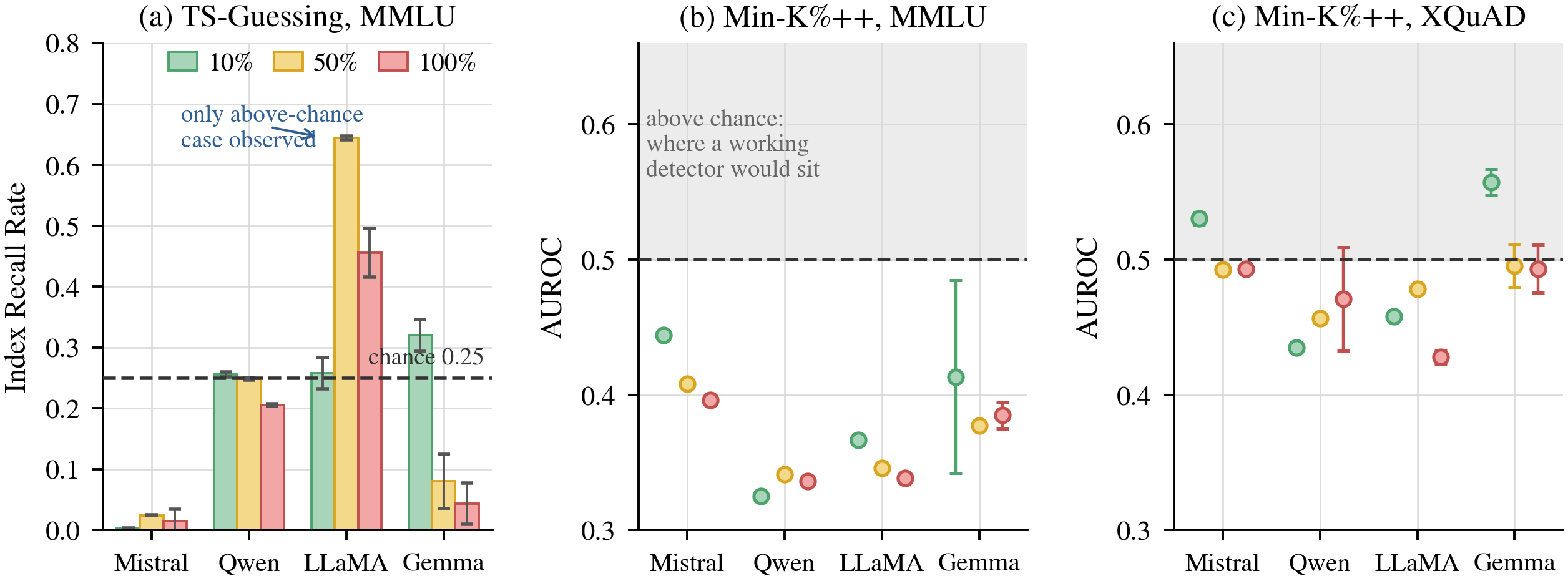}
  \caption{English-centric probes under translated exposure; error bars are standard deviations over three seeds. (a) TS-Guessing IDR on MMLU against the $0.25$ baseline: LLaMA at $50/100\%$ and Gemma at $10\%$ clearly exceed it, LLaMA and Qwen at $10\%$ are close to the baseline, and the remaining settings fall below it. (b, c) Min-K\%++ AUROC; the shaded band indicates the above-chance region. No MMLU setting enters this region, while two XQuAD settings do so marginally. Values are reported in Tables~\ref{tab:ts_side_by_side} and~\ref{tab:aucroc_results}.}
  \label{fig:probe_null}
\end{figure*}

\paragraph{4.2. English-centric probes provide limited evidence of the exposure effect.}
Figure~\ref{fig:probe_null}(a) shows that TS-Guessing provides a model-dependent signal. On MMLU, LLaMA is the clearest case: IDR rises from $0.258$ at $10\%$ exposure to $0.645$ at $50\%$, well above the $0.25$ random baseline, before decreasing to $0.456$ at $100\%$. Qwen remains close to the random baseline across exposure levels, never exceeding it by more than $0.006$ and decreasing to $0.206$ at full exposure. Mistral remains near zero, while Gemma decreases from $0.320$ at $10\%$ to $0.044$ at $100\%$. The variation around the $0.25$ reference indicates that TS-Guessing is sensitive to model-specific response behavior in addition to positional recall.

Masked-text recovery is also low and shows no consistent exposure trend (Appendix~\ref{app:full}, Table~\ref{tab:ts_side_by_side}). XQuAD is weaker still: EM remains below $0.07$ for every model and exposure level, with no common monotonic pattern.

These results distinguish downstream performance changes from their detectability through English-centric probes. Models improve on English MMLU after Arabic exposure even when targeted English-centric probes do not consistently recover the exposed content. The LLaMA positional-recall result shows that benchmark-specific associations can remain detectable under translated exposure, but this signal does not account for the broader cross-model performance pattern.

\begin{figure*}[t]
  \centering
  \includegraphics[width=\linewidth]{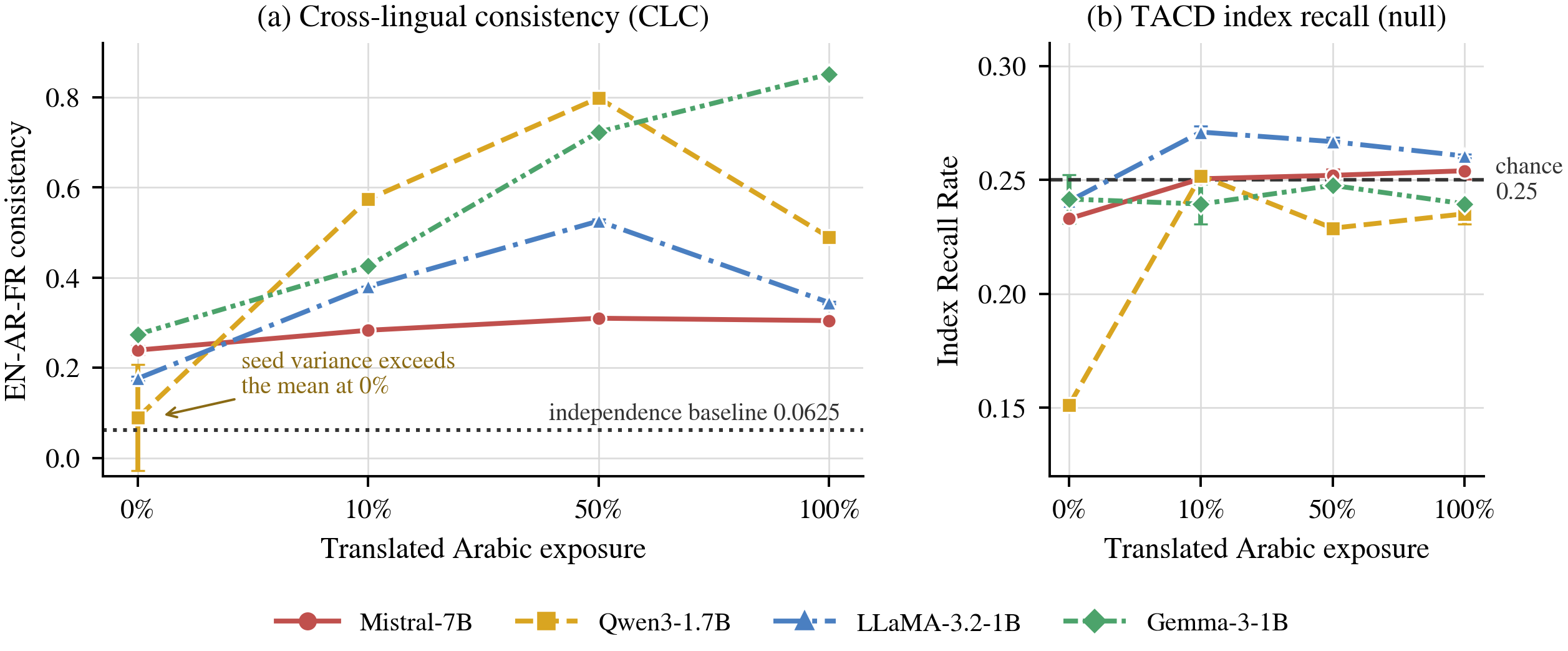}
  \caption{TACD signals; error bars are standard deviations over three seeds. (a) EN--AR--FR consistency is substantially above the $0.0625$ independence reference across models, with the largest changes from the clean condition observed for Qwen and Gemma. (b) Index recall remains close to the $0.25$ reference across conditions. Values are reported in Tables~\ref{tab:idr} and~\ref{tab:clc}.}
  \label{fig:tacd}
\end{figure*}

\paragraph{4.3. Min-K\%++ provides no above-chance contamination signal.}
Figure~\ref{fig:probe_null}(b, c) provides a complementary view based on token-level likelihoods. On MMLU, Min-K\%++ AUROC remains below $0.5$ for every model and exposure level, ranging from $0.33$ to $0.44$. LLaMA decreases from $0.367$ to $0.338$, while Mistral decreases from $0.444$ to $0.396$ between $10\%$ and $100\%$ exposure. On XQuAD, AUROC remains close to chance, ranging from $0.43$ to $0.56$, without a consistent exposure trend.

Because AUROC depends on the orientation of the member/non-member labels, values below $0.5$ are interpreted as the absence of the expected above-chance separation rather than as evidence of inverse detection. Overall, Min-K\%++ does not recover a consistent contamination signal under translated exposure.

\paragraph{4.4. Multilingual consistency provides a complementary signal.}
If translated exposure affects the model but English-only probes do not reliably reveal it, the relevant signal may lie in how the model behaves across languages. Figure~\ref{fig:tacd} reports the two TACD signals.

IDR remains close to the $0.25$ random baseline across models and exposure levels, while CLC exhibits substantial changes with translated exposure. Relative to each model's clean condition, CLC increases for LLaMA, Qwen, and Gemma: LLaMA rises from $0.176$ to a peak of $0.526$ at $50\%$ exposure, Qwen from $0.090$ to $0.799$ at $50\%$, and Gemma increases from $0.274$ to $0.851$ at full exposure. These values substantially exceed the $0.0625$ independence reference. Mistral also shows an increase, from $0.239$ in the clean condition to $0.310$ at $50\%$ and $0.305$ at $100\%$ exposure.

The results indicate that translated exposure can alter the degree of behavioral agreement across languages, with the magnitude and trajectory varying across models. Notably, LLaMA and Qwen reach their highest CLC at $50\%$ exposure, whereas Gemma continues to increase through full exposure. This variation shows that cross-lingual consistency captures a behavioral dimension that is distinct from English-only performance and contamination probes. In this controlled setting, TACD therefore provides complementary evidence of exposure-associated changes that are not consistently visible through English-centric evaluation alone.

\begin{figure*}[t]
  \centering
  \includegraphics[width=\linewidth]{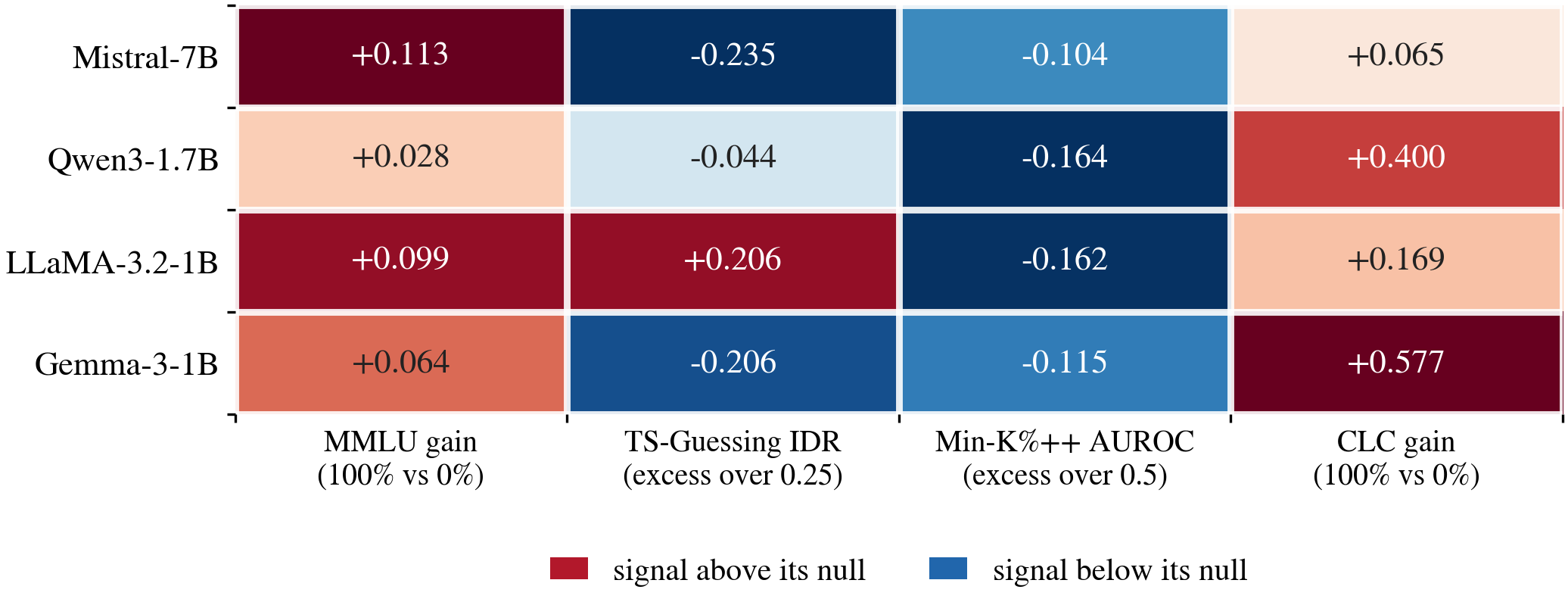}
  \caption{Every signal at full exposure as a deviation from its own reference: MMLU accuracy and CLC relative to the clean condition, IDR relative to $0.25$, and AUROC relative to $0.5$. Color is normalized within columns, so shades compare down but not across; raw deviations are printed. The signals exhibit distinct patterns across models, with performance changes and cross-lingual consistency not necessarily increasing in parallel.}
  \label{fig:signal_matrix}
\end{figure*}

\paragraph{4.5. Exposure effects vary across models.}
The four models exhibit distinct relationships between English performance and cross-lingual consistency. Mistral provides a particularly informative example: its English MMLU accuracy increases from $0.577$ to $0.690$, an absolute gain of $0.113$, while CLC increases from $0.239$ to $0.305$. Figure~\ref{fig:signal_matrix} shows that the different signals do not follow the same ordering across models. This demonstrates that translated exposure can affect performance and cross-lingual behavioral consistency through distinct patterns.

Taken together, the experiments show that translated exposure can produce measurable changes in English benchmark performance that are not consistently recovered by English-centric contamination probes. At the same time, cross-lingual behavioral consistency captures an additional exposure-associated signal, with its magnitude and trajectory depending on both the model and evaluation task. MMLU provides the clearest evidence for this relationship, while XQuAD shows that the same pattern does not automatically extend to context-rich extractive QA.

\section{Conclusion}
Benchmark contamination is difficult to audit when the exposed material differs from the evaluation benchmark in surface form. We studied this through a controlled translated-exposure setting in which four open-weight instruction-tuned LLMs received increasing amounts of Arabic benchmark data and were evaluated on the original English tasks.

The results establish a clear sequence on MMLU. Translated exposure increases English benchmark performance across all four models. The two English-centric probes we tested do not reliably recover this effect: TS-Guessing reveals strong positional recall only for LLaMA, while Min-K\%++ remains below chance. Comparing predictions across English, Arabic, and French reveals a complementary signal, with CLC substantially above an independence baseline and generally increasing relative to the clean condition for three of the four models. TACD formalizes this last comparison. It is not a definitive contamination detector or a membership test, but a training-data-free diagnostic for contamination-consistent cross-lingual behavior, and its model-dependent behavior, illustrated most clearly by Mistral, bounds how broadly the signal should be interpreted.

The broader lesson is that translation is not necessarily a safeguard against contamination. Changing the language of benchmark content can remove the surface correspondence on which conventional detectors depend while preserving enough semantic structure for the exposed content to affect evaluation. Contamination audits for multilingual benchmarks should therefore complement English-centric checks with multilingual behavioral analyses, and should distinguish evidence of cross-lingual invariance from definitive claims of training-set membership.

\section*{Limitations}

Our conclusions concern exposure-associated behavioral changes rather than definitive contamination detection. TACD measures cross-lingual consistency and does not establish training-set membership. Cross-lingual agreement may also arise from model-, prompt-, or translation-induced invariance. Future work should distinguish consistency on correct predictions from consistency on errors.

Our exposure conditions also vary total fine-tuning volume. Since $\mathcal{D}^{d}_{\text{AR}}(p)$ is added to a fixed English split, differences between $p=0$ and $p>0$ reflect both language and training volume. A stronger design would control total training volume and hold out evaluation items from the exposure set.

The contamination mechanism is controlled and artificial: we fine-tune on translated evaluation material using LoRA, whereas natural contamination may arise through pretraining or continued training with partial, paraphrased, or noisy exposure. The results therefore characterize a controlled exposure mechanism rather than naturally occurring contamination.

Finally, the study covers four open-weight models, two benchmarks, and three languages. The observed patterns vary across models and tasks, with MMLU showing the clearest effects and XQuAD showing weaker transfer. Translation quality and seed variance introduce additional variation, particularly for the clean Qwen condition. Min-K\%++ is interpreted relative to the $0.5$ chance level because no clean-reference run was included.

\appendix
\section{Full Numerical Results}
\label{app:full}

This appendix reports the numerical values underlying Figures~\ref{fig:exposure_curves}--\ref{fig:signal_matrix}. Probe and diagnostic results are means over three random seeds ($\{1337,42,99\}$) with standard deviations. Model names are abbreviated after Table~\ref{tab:mmlu_xquad_per_model}.

\begin{table}[H]
\centering
\scriptsize
\setlength{\tabcolsep}{3pt}
\renewcommand{\arraystretch}{1.05}
\caption{IDR across translated-exposure levels (mean~$\pm$~std). Random baseline $=0.25$.}
\label{tab:idr}
\begin{tabular}{lcccc}
\toprule
Model & 0\% & 10\% & 50\% & 100\% \\
\midrule
LLaMA   & .241$\pm$.003 & .271$\pm$.003 & .267$\pm$.002 & .261$\pm$.001 \\
Qwen    & .151$\pm$.001 & .252$\pm$.002 & .229$\pm$.001 & .235$\pm$.005 \\
Gemma   & .242$\pm$.011 & .239$\pm$.009 & .248$\pm$.001 & .240$\pm$.001 \\
Mistral & .233$\pm$.002 & .251$\pm$.002 & .252$\pm$.003 & .254$\pm$.004 \\
\bottomrule
\end{tabular}
\end{table}

\begin{table}[H]
\centering
\scriptsize
\setlength{\tabcolsep}{3pt}
\renewcommand{\arraystretch}{1.05}
\caption{CLC across translated-exposure levels (mean~$\pm$~std). Independence baseline $\approx0.0625$.}
\label{tab:clc}
\begin{tabular}{lcccc}
\toprule
Model & 0\% & 10\% & 50\% & 100\% \\
\midrule
LLaMA   & .176$\pm$.005 & .380$\pm$.000 & .526$\pm$.004 & .345$\pm$.003 \\
Qwen    & .090$\pm$.117 & .574$\pm$.001 & .799$\pm$.001 & .490$\pm$.000 \\
Gemma   & .274$\pm$.004 & .426$\pm$.003 & .723$\pm$.006 & .851$\pm$.007 \\
Mistral & .239$\pm$.009 & .283$\pm$.001 & .310$\pm$.002 & .305$\pm$.006 \\
\bottomrule
\end{tabular}
\end{table}

\begin{table}[H]
\centering
\footnotesize
\setlength{\tabcolsep}{5pt}
\caption{English MMLU and XQuAD with the LM Evaluation Harness, by translated-exposure level. Plotted in Figure~\ref{fig:exposure_curves}.}
\label{tab:mmlu_xquad_per_model}
\begin{tabular}{lcccc}
\toprule
Model & 0\% & 10\% & 50\% & 100\% \\
\midrule
\multicolumn{5}{l}{MMLU, macro-averaged accuracy} \\
Mistral-7B-Instruct-v0.2 & 0.577 & 0.580 & 0.600 & 0.690 \\
Qwen3-1.7B               & 0.553 & 0.560 & 0.562 & 0.581 \\
LLaMA-3.2-1B-Instruct    & 0.332 & 0.381 & 0.389 & 0.431 \\
Gemma-3-1B-it            & 0.220 & 0.244 & 0.261 & 0.284 \\
\midrule
\multicolumn{5}{l}{XQuAD, ROUGE-L F1} \\
Mistral-7B-Instruct-v0.2 & 0.302 & 0.455 & 0.272 & 0.114 \\
Qwen3-1.7B               & 0.457 & 0.429 & 0.510 & 0.564 \\
LLaMA-3.2-1B-Instruct    & 0.364 & 0.459 & 0.558 & 0.569 \\
Gemma-3-1B-it            & 0.364 & 0.481 & 0.577 & 0.606 \\
\bottomrule
\end{tabular}
\end{table}

\begin{table}[H]
\centering
\scriptsize
\setlength{\tabcolsep}{2pt}
\caption{TS-Guessing under translated exposure, mean~$\pm$~std over three seeds. The MMLU index-recall block is plotted in Figure~\ref{fig:probe_null}(a).}
\label{tab:ts_side_by_side}
\begin{tabular}{lccc}
\toprule
Model & 10\% & 50\% & 100\% \\
\midrule
\multicolumn{4}{l}{MMLU, index recall (IDR); random baseline $0.25$} \\
LLaMA   & $0.2577\pm0.0254$ & $0.6447\pm0.0029$ & $0.4560\pm0.0399$ \\
Qwen    & $0.2560\pm0.0044$ & $0.2487\pm0.0021$ & $0.2060\pm0.0020$ \\
Gemma   & $0.3200\pm0.0260$ & $0.0803\pm0.0446$ & $0.0437\pm0.0335$ \\
Mistral & $0.0021\pm0.0019$ & $0.0245\pm0.0007$ & $0.0150\pm0.0190$ \\
\midrule
\multicolumn{4}{l}{MMLU, masked-choice recovery (RL-F1)} \\
LLaMA   & $0.0401\pm0.0533$ & $0.0038\pm0.0019$ & $0.0036\pm0.0021$ \\
Qwen    & $0.0010\pm0.0000$ & $0.0188\pm0.0003$ & $0.0140\pm0.0002$ \\
Gemma   & $0.0604\pm0.0056$ & $0.1340\pm0.0165$ & $0.1169\pm0.0501$ \\
Mistral & $0.0804\pm0.0358$ & $0.0714\pm0.0246$ & $0.0765\pm0.0342$ \\
\midrule
\multicolumn{4}{l}{XQuAD, masked-token recovery (EM)} \\
LLaMA   & $0.0403\pm0.0597$ & $0.0090\pm0.0035$ & $0.0058\pm0.0020$ \\
Qwen    & $0.0003\pm0.0005$ & $0.0089\pm0.0102$ & $0.0328\pm0.0258$ \\
Gemma   & $0.0169\pm0.0025$ & $0.0195\pm0.0066$ & $0.0101\pm0.0045$ \\
Mistral & $0.0692\pm0.0294$ & $0.0682\pm0.0215$ & $0.0459\pm0.0247$ \\
\midrule
\multicolumn{4}{l}{XQuAD, masked-token recovery (RL-F1)} \\
LLaMA   & $0.0096\pm0.0067$ & $0.0109\pm0.0051$ & $0.0077\pm0.0020$ \\
Qwen    & $0.0003\pm0.0005$ & $0.0104\pm0.0112$ & $0.0375\pm0.0299$ \\
Gemma   & $0.0194\pm0.0036$ & $0.0215\pm0.0077$ & $0.0107\pm0.0033$ \\
Mistral & $0.0819\pm0.0270$ & $0.0770\pm0.0208$ & $0.0551\pm0.0247$ \\
\bottomrule
\end{tabular}
\end{table}

\begin{table}[H]
\centering
\scriptsize
\setlength{\tabcolsep}{2pt}
\caption{Min-K\%++ AUROC under translated exposure, mean~$\pm$~std over three seeds. Chance is $0.5$; no clean-reference run is available. Plotted in Figure~\ref{fig:probe_null}(b, c).}
\label{tab:aucroc_results}
\begin{tabular}{lccc}
\toprule
Model & 10\% & 50\% & 100\% \\
\midrule
\multicolumn{4}{l}{MMLU} \\
LLaMA   & $0.3667\pm0.0015$ & $0.3460\pm0.0035$ & $0.3383\pm0.0015$ \\
Qwen    & $0.3253\pm0.0006$ & $0.3413\pm0.0012$ & $0.3360\pm0.0010$ \\
Gemma   & $0.4133\pm0.0713$ & $0.3774\pm0.0006$ & $0.3850\pm0.0098$ \\
Mistral & $0.4440\pm0.0010$ & $0.4080\pm0.0010$ & $0.3960\pm0.0010$ \\
\midrule
\multicolumn{4}{l}{XQuAD} \\
LLaMA   & $0.4580\pm0.0009$ & $0.4781\pm0.0016$ & $0.4279\pm0.0051$ \\
Qwen    & $0.4350\pm0.0035$ & $0.4568\pm0.0007$ & $0.4709\pm0.0382$ \\
Gemma   & $0.5573\pm0.0098$ & $0.4956\pm0.0161$ & $0.4933\pm0.0176$ \\
Mistral & $0.5303\pm0.0046$ & $0.4924\pm0.0012$ & $0.4930\pm0.0009$ \\
\bottomrule
\end{tabular}
\end{table}

\bibliographystyle{acl_natbib}
\bibliography{tacl2021}

\end{document}